\documentclass[10pt, a4paper]{article}
\usepackage{lrec}
\usepackage{graphicx}
\usepackage{tabularx}
\usepackage{color}
\usepackage{soul}
\usepackage{float}
\usepackage{epstopdf}
\usepackage[latin1]{inputenc}
\usepackage{hyperref}
\usepackage{xstring}
\usepackage{titlesec}
\titleformat{\section}{\normalfont\large\bf\center}{\thesection.}{1em}{}
\titleformat{\subsection}{\normalfont\SmallTitleFont\bf\raggedright}{\thesubsection.}{1em}{}
\titleformat{\subsubsection}{\normalfont\normalsize\bf\raggedright}{\thesubsubsection.}{1em}{}
\renewcommand\thesection{\arabic{section}}
\renewcommand\thesubsection{\thesection.\arabic{subsection}}
\renewcommand\thesubsubsection{\thesubsection.\arabic{subsubsection}}

\title{Multilingual Culture-Independent Word Analogy Datasets}

\name{Matej Ul\v{c}ar$^{1}$, Kristiina Vaik$^{2,4}$, Jessica Lindstr\"{o}m$^{3}$, Milda Dailid\.{e}nait\.{e}$^{4,5}$, Marko Robnik-\v{S}ikonja$^{1}$}

\address{$^1$University of Ljubljana, Faculty of Computer and Information Science, \\
         $^2$Texta O\"{U}, Tallinn, Estonia,
         $^3$University of Helsinki, Department of Finnish, Finno-Ugrian and Scandinavian Studies, \\
         $^4$University of Tartu, Faculty of Arts and Humanities,
         $^5$University of Latvia, Livonian institute, \\
         $^1$Ve\v{c}na pot 113, SI-1000 Ljubljana, Slovenia\\
         \{matej.ulcar, marko.robnik\}@fri.uni-lj.si, kristiina.vaik@ut.ee, jessica.lindstrom@helsinki.fi, m.dailidenaite@gmail.com}

\abstract{
In text processing, deep neural networks mostly use word embeddings as an input. Embeddings have to ensure that relations between words are reflected through distances in a high-dimensional numeric space.  To compare the quality of different text embeddings, typically, we use benchmark datasets.  We present a collection of such datasets for the word analogy task in nine languages:	Croatian, English, 	Estonian, Finnish, Latvian, Lithuanian, Russian,  Slovenian, and Swedish. We designed the monolingual analogy task to be much more culturally independent and also constructed cross-lingual analogy datasets for the involved languages.
We present basic statistics of the created datasets and their initial evaluation using fastText embeddings. 
 \\ \newline  }

\begin{document}

\maketitleabstract

\section{Introduction}
As an input, neural networks require numerical data. \emph{Text embeddings}  provide such an input, ensuring that relations between words are reflected in the distances and directions in high-dimensional numeric space.  There are many distinct models producing embedding vectors, using different specialized learning tasks, e.g., word2vec \cite{mikolov2013exploiting}, GloVe \cite{pennington2014glove}, and fastText \cite{Bojanowski2017}.  For training, the embeddings algorithms use  large monolingual corpora. 


To compare the quality of different text embeddings, typically we use benchmark datasets.  In this work, we present a collection of such datasets for the word analogy task in nine languages: Croatian, English, Estonian, Finnish, Latvian, Lithuanian, Russian,  Slovenian, and Swedish. To make the datasets sensible for all languages, we designed the analogy task to be culturally neutral. Specifically, we avoided analogy categories that exhibit single culture or single country examples, such as NFL teams, US cities, or US states.

The word analogy  task was  popularized  by \newcite{mikolov2013distributed}.   The goal is to find a term $y$ for a given term $x$ so that the relationship between $x$ and $y$ best resembles the given relationship $a : b$.
There are two main groups of categories: semantic and syntactic. To illustrate a semantic relationship, consider for example that the word pair $a : b$ is given as ``Finland : Helsinki''. The task is to find the term $y$ corresponding to the relationship ``Sweden : $y$'', with the expected answer being $y=$ Stockholm. In syntactic categories, each category refers to a grammatical feature, for example adjective degrees of comparison. The two words in any given pair then have a common stem (or even the same lemma), for example, given the word pair ``long : longer'', we see that we have an adjective in its base form and the same adjective in a comparative form. The task is then to find the term $y$ corresponding to the relationship ``dark : $y$'', with the expected answer being $y=$ darker, i.e. a comparative form of the adjective dark. 

In the vector space, the analogy task is transformed into vector arithmetic and we search for nearest neighbours, i.e. we compute the distance between vectors: d(vec(Finland), vec(Helsinki)) and search for word $y$ which would give the closest result in distance  d(vec(Sweden), vec($y$)). 
In our dataset, the analogies are already prespecified, so we do not search for the closest result but only check if the prespecified word is indeed the closest, or alternatively, measure the distance between the given pairs.

The paper is split into further four sections. In Section \ref{sec:analogy}, we describe the analogy task, its origin, culture-independent design, structure, and how it can be used as a benchmark for evaluation of embeddings in monolingual and cross-lingual setting. In Section \ref{sec:datasets}, we present the creation of the actual monolingual and cross-lingual datasets and the process of their adaptation to all involved languages. We present statistics and initial evaluations of the produced datasets  in Section \ref{sec:evaluation}.
Conclusion and plans for further work are described in Section   \ref{sec:conclusions}.

\section{Analogy task for embedding evaluations}
\label{sec:analogy}
We composed the analogy tasks for nine languages from the EMBEDDIA project\footnote{EMBEDDIA: Cross-Lingual Embeddings for Less-Represented Languages in European News Media, \url{http://embeddia.eu}}. The work is based on the English dataset by \newcite{mikolov2013efficient}\footnote{\url{http://download.tensorflow.org/data/questions-words.txt}}. Due to English- and US-centered bias of this dataset, we removed some categories and added or changed some of the others as described below. Our dataset was first written in Slovene language and then translated to other languages as explained in Section \ref{sec:translation}. Following  \newcite{mikolov2013efficient}, we limit the analogies to single word terms; for example ``New Zealand'' is not used as a term for a country, since it consists of two words. Note that due to language differences, the produced datasets are not aligned across languages.

To assure consistency and allow the use of the datasets in cross-lingual analogies (described in Section \ref{sec:cross-lingual-analogies}), our datasets (even the English one) are somewhat different from the one by \newcite{mikolov2013efficient}. We removed or edited some categories and added new ones to avoid or limit English-centric bias in the following way.
\begin{itemize}
    \item We merged two categories dealing with countries and their capitals ("common capital cities" and "all capital cities") into one category.
    \item We changed "city in US state" category to "city in country" and used mostly European countries with a better chance to appear in the corpora of respective languages. 
    \item We removed the category "currency", as only a handful of currencies are nowadays present in news and text corpora with sufficient frequency. 
    \item We added two new semantic categories, "animals" and "city with river" described below. 
    \item We added a syntactic category comparing noun case relationships. 
\end{itemize}

The resulting analogy tasks are composed of 15 categories: 5 semantic and 10 syntactic/morphological. The categories contained in our datasets are the following:
\begin{description}
\item [capitals and countries,] capital cities in relation to countries, e.g., Paris : France,
\item [family,] a male family member in relation to an equivalent female member, e.g., brother : sister,
\item [city in country,] a non-capital city in relation to the country of that city, e.g., Frankfurt : Germany,
\item [animals,] species/subspecies in relation to their genus/familia, following colloquial terminology and relations, not scientific, e.g., salmon : fish,
\item [city with river,] a city in relation to the river flowing through it, e.g., London : Thames,
\item [adjective to adverb,] an adverb in relation to the adjective it is formed from, e.g., quiet : quietly,
\item [opposite adjective,] the morphologically derived opposite adjective in relation to the base form, e.g., just : unjust, or honest : dishonest,
\item [comparative adjective,]  the comparative form of adjective in relation to the base form, e.g., long : longer,
\item [superlative adjective,]  the superlative form of adjective in relation to the base form, e.g., long : longest,
\item [verb to verbal noun,] noun formed from verb in infinitive form, e.g., to sit : sitting; in Estonian and Finnish -da infinitive and first infinitive forms are used respectively; in Swedish present participle that functions as noun is used in place of verbal noun,
\item [country to nationality] of its inhabitants, e.g., Albania : Albanians,
\item [singular to plural,] singular form of a noun in relation to the plural form of the noun, e.g., computer : computers; indefinite singular and definite plural are used in Swedish,
\item [genitive to dative,] a genitive noun case in relation to the dative noun case in respective languages, e.g. in Slovene ceste : cesti: singular is used for all words, except "human" (or equivalent in other languages), which appears in both singular and plural; in Finnish and Estonian, dative has been replaced with the allative case; the category is not applicable to Swedish and English,
\item [present to past,] 3rd person singular verb in present tense in relation to 3rd person singular verb in past tense, e.g., goes : went; in Slovene, Croatian, and Russian the masculine gender past tense is used, in other languages the "simple" past tense/preterite is used,
\item [present to other tense,] 3rd person singular verb in present tense in relation to the 3rd person singular verb in various tenses, e.g., goes : gone; the other tense in Slovene, Croatian, and Russian is feminine gender past tense; in Finnish, Estonian, and English it is present/past perfect participle; in Swedish it is supine; in Latvian and Lithuanian it is future tense.
\end{description}

\subsection{Cross-lingual analogies}
\label{sec:cross-lingual-analogies}
Cross-lingual word embeddings have two or more languages in the same semantic vector space. Cross-lingual word analogy task has been proposed by \newcite{brychcin2019cross} as an intrinsic evaluation of cross-lingual embeddings. Following their work, we compose cross-lingual analogy datasets, so that one pair of related words is in one language and the other pair from the same category is in another language. For example, given the relationship in English father : mother, the task is to find the term $y$ corresponding to the relationship brat (brother) : $y$ in Slovene. The expected answer being $y =$ sestra (sister). We limited the cross-lingual analogies to the categories that all our languages have in common, i.e. we excluded the last three syntactic categories: genitive to dative, present to past, and present to other tense. 

\section{Creation of datasets}
\label{sec:datasets}
Once the relations forming the analogies were prepared, we used them to form the actual monolingual and cross-lingual datasets. The process consisted of three steps. In Sections \ref{sec:monolingual} and \ref{sec:crosslingual}, we describe the creation of monolingual and cross-lingual datasets from the relations, and in Section \ref{sec:translation}, we explain the translation procedure which lead to creation of 
datasets in all involved languages.

\subsection{Compiling monolingual dataset}
\label{sec:monolingual}
The actual construction of the analogy dataset started by forming baseline relations for each category. 
First, we manually wrote the relations one per line, where each relation consists of two words. In the family category, an example of such a relation is ``father, mother''. Next we combined all relations in each category with one another and wrote them in pairs, e.g., ``father, mother, brother, sister'' If a pair of relations share a common word, such a pair is excluded from the database. An example of forming relation pairs is shown in Table \ref{tab:compiling}.

\begin{table}[!h]
\begin{center}
\begin{tabularx}{0.8\columnwidth}{Xlll}
		Relations & & &  \\ \hline
		Vienna & Danube & & \\
		Budapest & Danube & & \\
		Cairo & Nile & & \\
		Paris & Seine & & \\ \hline
		& & & \\
		Formed pairs & & & \\ \hline
		Vienna & Danube & Cairo & Nile \\
        Vienna & Danube & Paris & Seine \\
		Budapest & Danube & Cairo & Nile \\
		Budapest & Danube & Paris & Seine \\
		Cairo & Nile & Paris & Seine \\
      \hline
\end{tabularx}
\caption{An excerpt from the ``city with river'' category, showing four relations and five relation pairs formed from them. The first two listed relations do not form a pair with each other, because they share a common word (Danube).}
\label{tab:compiling}
 \end{center}
\end{table}

\subsection{Cross-lingual datasets}
\label{sec:crosslingual}
Cross-lingual analogies described in Section \ref{sec:cross-lingual-analogies} are compiled in a similar manner. Consider a language pair $L_1 - L_2$. From the same one-relation-per-line files shown in the upper part of Table  \ref{tab:compiling}, we combine all relations in a category in such a way that one relation from language $L_1$ and one relation from language $L_2$ form a pair. $L_1$ relations are on the left-hand side, and $L_2$ relations are on the right-hand side. An example of forming cross-lingual relation pairs is shown in Table \ref{tab:xlcompiling} for English-Slovene language pair. The same rules for excluding pairs with common words apply, except that we do not consider translations of the same term as the same word, e.g., ``Nile'' (in English) and ``Nil'' (its Slovene equivalent) in the same entry are allowed, but using ``Nile'' twice is disallowed. 

\begin{table}[!h]
\begin{center}
\begin{tabularx}{0.9\columnwidth}{llll}
		\multicolumn{4}{l}{Relations: English} \\ \hline
		Vienna & Danube & & \\
		Budapest & Danube & & \\
		\hline
		& & & \\
		\multicolumn{4}{l}{Relations: Slovene} \\ \hline
		Budimpe\v{s}ta & Donava & & \\
		Kairo & Nil & & \\ \hline
		& & & \\
		\multicolumn{4}{l}{Formed pairs (English-Slovene)}  \\ \hline
		Vienna & Danube & Budimpe\v{s}ta & Donava \\
        Vienna & Danube & Kairo & Nil \\
		Budapest & Danube & Budimpe\v{s}ta & Donava \\
		Budapest & Danube & Kairo & Nil \\
      \hline
\end{tabularx}
\caption{An excerpt from the ``city with river'' category, showing two relations in English, two relations in Slovene and four relation pairs formed from them in a crosslingual English-Slovene analogy dataset.}
 \label{tab:xlcompiling}
 \end{center}
\end{table}

\subsection{Translation procedure}
\label{sec:translation}
When the first dataset in Slovene was formed, we translated it into other languages (including English).
We used various tools to help us translate Slovenian dataset to the other languages. For the geographic data, i.e. names of countries, cities and rivers, we used the titles of equivalent Wikipedia articles or data from Wikipedia lists, such as the list of capital cities. If an entity had a name consisting of more than one word in another language, it was either skipped or replaced by another entity with subjectively similar location and/or importance. The same was done in cases where we would have a relation of type ``x : x'', which is nonsensical. For example, in Lithuanian Algeria and its capital Algiers are both called ``Al\v{z}yras''. So we replaced it with ``Damaskas : Sirija'' (in English this would  correspond to ``Damascus : Syria'').

For non-geographic words, we mostly used Babelnet\footnote{\url{https://babelnet.org/}} and Wiktionary\footnote{\url{https://wiktionary.org}} to find the translations. In the latter, we mostly relied on conjugation and declination tables of our key words. Wiktionary was also used for finding new examples for relations in syntactical categories, to replace those for which a translation was either impossible or could not be found. This was most often the case in all the categories operating with adjectives. An example of an impossible translation is the Slovene relation  "drag : dra\v{z}ji". Its English translation is "expensive : more expensive". Since we are limited to single-word terms, we discarded that translation and replaced such a relation with another one, with either a similar meaning "costly : costlier", or a completely different one, like "high : higher", provided it does not already appear in the dataset.

English and Swedish languages do not have noun cases or rather only have genitive case (in addition to nominative) in a very limited sense. We decided to exclude the ``genitive to dative'' category for these two languages. Further more, while Finnish and Estonian have many noun cases, none of those cases is dative. We exchanged dative in this category with allative case, which mostly covers the same role. 

For two categories, ``city with river'' and in a smaller part ``city in country'' we intentionally varied the entries across languages more than in other categories, where it was only done so out of necessity. We felt certain relations are too locally specific to frequently (or at all) appear in other language corpora. We removed most of such relations in other languages and tried to replace them with other relations more geographically local to that language, in order to keep the number of different countries or rivers high. Majority of the relations in these two categories is still the same for all languages.  

The translated relations were checked by native speakers of each language and corrected where deemed necessary. 

\section{Statistics and evaluation}
\label{sec:evaluation}
In this section, we first present relevant statistics of the created datasets, followed by their evaluation using fastText embeddings.

\subsection{Statistics}
The original English analogy dataset by \newcite{mikolov2013efficient} contains 19,544 relations, but uses slightly different categories to our datasets. As explained above, we translated the Slovene dataset into all other languages to keep datasets similar across languages, especially for the use in cross-lingual analogy tasks.
The number of obtained analogy pairs in monolingual datasets is between 18,000 to 20,000 per language. The exact numbers differ from language to language based on the validity of categories and availability of sensible examples in each category. The exact numbers for all languages are shown in Table \ref{tab:analogy}. 

\begin{table}[!h]
\begin{center}
\begin{tabularx}{0.5\columnwidth}{Xr}
		Language	& Size  \\ \hline
		Croatian	&  19416 \\ 
		English  	&  18530 \\
		Estonian	&  18372 \\ 
		Finnish  	&  19462 \\ 
		Latvian  	&  20138 \\ 
		Lithuanian	&  20022 \\ 
		Russian 	&  19976 \\ 
		Slovene 	&  19918 \\ 
		Swedish	    &  18480 \\ 
      \hline
\end{tabularx}
\caption{The sizes of the constructed monolingual word analogy datasets expressed as numbers of pairs for each language.}
 \label{tab:analogy}
 \end{center}
\end{table}

The number of pairs in cross-lingual datasets is smaller, because some categories were omitted. We created cross-lingual datasets for all 72 language pairs. The exact sizes of datasets for a few selected pairs are shown in Table \ref{tab:xlsize}.

\begin{table}[!h]
\begin{center}
\begin{tabularx}{0.8\columnwidth}{Xr}
		Language pair	& Size  \\ \hline
		Croatian-English & 17667 \\
		Croatian-Slovene  	&  17449 \\
		English-Slovene	&  17964 \\
		Estonian-Finnish & 16809 \\
		Estonian-Slovene	&  17110 \\
		Finnish-Swedish & 17600 \\
		Latvian-Lithuanian & 18056 \\
      \hline
\end{tabularx}
\caption{The sizes of a few constructed cross-lingual word analogy datasets expressed as numbers of pairs for each language.  }
 \label{tab:xlsize}
 \end{center}
\end{table}

Not all categories are equally represented, some have much more relation pairs than others. We tried to downplay the importance of the category ``capitals and countries'', which is very prominent in the dataset by \newcite{mikolov2013efficient}, however, it is still by far the largest category in our dataset. Some categories are necessarily small, like ``family'', since the number of terms for family members is relatively small. That is especially true for languages from northern Europe, so we also included plural terms and some non-family members in that category, like a relation ``king : queen''. The number of analogy pairs per category (averaged over all languages) is shown in Table \ref{tab:catsize}. Due to this difference in sizes, we strongly suggest that results are presented for each category separately, or when aggregated, to report the macro average score (average of category scores, not average over all).

\begin{table}[!h]
\begin{center}

\begin{tabularx}{0.9\columnwidth}{Xr}
		Category	& Average size  \\ \hline
		Capitals and countries	&  5701 \\ 
		Family  	&  482 \\
		City in country	&  2880 \\ 
		Animals  	&  1440 \\ 
		City with river  	&  701 \\ 
		Adjective to adverb	&  873 \\ 
		Opposite adjective	&  498 \\ 
		Comparative adjective	&  866 \\ 
		Superlative	adjective    &  823 \\ 
		Verb to verbal noun & 415 \\
		Country to nationality & 924 \\
		Singular to plural & 1519 \\
		Genitive to dative & 1356 \\
		Present to past & 607 \\
		Present to other tense & 601 \\
      \hline
\end{tabularx}
\caption{Average size in number of pairs for each category in the monolingual word analogy datasets.}
 \label{tab:catsize}
 \end{center}
\end{table}

\subsection{Evaluation}

We evaluated the analogy datasets using the fastText \cite{Bojanowski2017} embeddings\footnote{\url{https://fasttext.cc/}}. The fastText embeddings use subword inputs which are suitable also for morphologically rich languages, we processed. We limited the evaluation to the word vectors of the 200,000 most frequent tokens from the embeddings of each language. Not all analogy pairs can be evaluated in that way, since some words do not appear among the first 200,000 words. The amount of pairs that are covered (i.e. all four words from the analogy are among the most frequent 200,000 words) for each language is shown in the Table \ref{tab:coverage}. 
\begin{table}[!h]
\begin{center}
\begin{tabularx}{0.6\columnwidth}{Xr}
		Language	& Coverage (\%)  \\ \hline
		Croatian	&  81.67 \\ 
		English  	&  97.05 \\
		Estonian	&  82.56 \\ 
		Finnish  	&  63.97 \\ 
		Latvian  	&  73.60 \\ 
		Lithuanian	&  77.66 \\ 
		Russian 	&  62.53 \\ 
		Slovene 	&  86.70 \\ 
		Swedish	    &  82.44 \\ 
      \hline
\end{tabularx}
\caption{Percentage of constructed analogy pairs covered by the first 200,000 word vectors from common crawl fastText embeddings.}
 \label{tab:coverage}
 \end{center}
\end{table}

\begin{table*}[!!htb]
\begin{center}
\begin{tabular}{lrrrrrrrrr}
Category & sl & en & hr & et & fi & lv & lt & sv & ru \\ \hline 
Capitals and countries & 28.13 & 95.23 & 37.11 & 43.33 & 79.09 & 45.90 & 53.75 & 88.38 & 81.26  \\
Family & 38.77 & 92.03 & 44.58 & 48.79 & 62.67 & 44.50 & 54.78 & 68.10 & 58.64  \\
City in country & 45.44 & 89.92 & 47.21 & 46.34 & 85.31 & 56.66 & 63.25 & 88.38 & 95.26  \\
Animals & 1.13 & 11.72 & 0.85 & 0.52 & 18.24 & 1.93 & 1.26 & 10.88 & 14.90  \\
City with river & 5.92 & 44.81 & 3.21 & 8.45 & 9.46 & 2.33 & 6.30 & 28.5 & 11.34  \\
Adjective to adverb & 36.62 & 27.32 & 34.76 & 48.40 & 53.66 & 53.22 & 60.58 & 84.33 & 29.31  \\
Opposite adjective & 30.42 & 50.00 & 33.01 & 38.60 & 24.74 & 36.36 & 55.00 & 16.14 & 0.00  \\
Comparative adjective & 31.38 & 96.88 & 36.40 & 72.36 & 75.69 & 68.65 & 55.03 & 78.82 & 37.55  \\
Superlative adjective & 19.28 & 97.31 & 18.03 & 28.07 & 59.47 & 10.04 & 52.84 & 38.31 & 23.08  \\
Verb to verbal noun & 65.33 & 82.37 & 59.76 & 93.27 & 86.25 & 65.68 & 58.82 & 31.85 & 19.05  \\
Country to nationality & 31.43 & 56.56 & 48.69 & 43.60 & 53.45 & 35.06 & 46.67 & 70.86 & 67.71  \\
Singular to plural & 32.68 & 91.78 & 34.89 & 72.16 & 87.16 & 42.38 & 51.13 & 41.23 & 57.35  \\
Genitive to dative & 26.68 & N/A & 31.76 & 61.29 & 46.48 & 39.91 & 22.44 & N/A & 33.19  \\
Present to past & 51.63 & 76.50 & 63.02 & 90.50 & 86.36 & 79.17 & 68.58 & 89.13 & 77.00  \\
Present to other tense & 54.17 & 32.55 & 54.07 & 69.83 & 82.64 & 62.94 & 61.90 & 87.15 & 78.50  \\
\hline
\end{tabular}
\caption{FastText evaluation scores in \% of correctly predicted relation pairs, i.e. how often was the vector $d$ the closest to the vector $b-a+c$, given a relation pair $a:b~ \approx ~c:d$. }
\label{tab:scores}
\end{center}
\end{table*}

We evaluated the relations that are completely contained in the first 200,000 fastText vectors. Given a pair of relations ``a : b $\approx$ c : d'', we searched for the closest word vector to the vector $b-a+c$, using cosine distance metric. In our search, we excluded the vectors $a$, $b$ and $c$. We report the number of times the closest word vector was vector of the word $d$. The results for all languages per category are shown in Table \ref{tab:scores}.

The results show that not all relations are recognized with the same accuracy across languages, the differences being large and surprising in some cases. This hints that there is a considerable space for improvement in construction of word embeddings.

\subsubsection{Error analysis}
We took a closer look at some of the categories with the largest differences in evaluation scores between Slovenian and English. The differences are on average the largest between these two languages.

The category ``Capitals and countries'' has an excellent score in English, but in Slovenian, the predictions face three issues: country names endings, morphological richness of the language and generally weak word embeddings, possibly due to low frequency of certain words considered in the analogy dataset. As the top part of Table \ref{tab:error-capital} demonstrates. some country names in Slovenian are homonymous to that country's adjectival form ("-ska" and "-\v{s}ka" endings"). When we search for the word vector, closest to $d$, we often get the adjectival form of country $d$ instead of the country's name. 

Another frequent error is that we get the same word as $c$, just in a different case (bottom part of Table \ref{tab:error-capital}). Finnish, also a morphologically rich language, scores much better in this category. Although, it does not seem to suffer from the first identified issue in Slovenian, it does suffer from the second issue, just much less frequently. This leads us to believe, that the problems identified are avoidable with even larger datasets and better quality of embeddings.

\begin{table}[htb]
\begin{center}
\begin{tabularx}{1.\columnwidth}{l|l}
		\multicolumn{2}{l}{Issue: adjectival form} \\ \hline
		Oslo Norve\v{s}ka  Canberra  Avstralija & Avstralska \\
		Oslo Norve\v{s}ka  Havana  Kuba & Kubanska \\
		Var\v{s}ava Poljska  Manila  Filipini & filipinska \\
		Stockholm \v{S}vedska  Tirana  Albanija & albanska \\
		\hline
		\multicolumn{2}{l}{} \\
		\multicolumn{2}{l}{Issue: cases} \\ \hline
		Stockholm \v{S}vedska  Bejrut  Libanon & Bejrutu \\
		Harare Zimbabve  Tirana  Albanija & Tirani \\
		Harare Zimbabve  Canberra  Avstralija & Canberri \\
		Nikozija Ciper  Stockholm  \v{S}vedska & Stockholma \\
      \hline
\end{tabularx}
\caption{Examples of two frequent errors in Slovenian in the ``Capitals and countries'' category. The four words one the left represent words a, b, c and d, respectively, from a dataset entry. In the fifth column is the best prediction of the word from the third column, as described in Section~\ref{sec:evaluation}.}
 \label{tab:error-capital}
 \end{center}
\end{table}

\begin{table}[htb]
\begin{center}
\begin{tabularx}{0.8\columnwidth}{lll}
        $c$ & $d$ & prediction \\ \hline
		trda & najtr\v{s}a & mehka \\
		hard & hardest & soft \\ \hline
		zahtevno & najzahtevnej\v{s}e & nezahtevno \\
		difficult & most difficult & non-difficult \\ \hline
		temno & najtemnej\v{s}e & svetlo \\
		dark & darkest & bright \\ \hline
		slab & najslab\v{s}i & dober \\
		bad & worst & good \\ \hline
		mlada & najmlaj\v{s}a & premlada \\
		young & youngest & too young \\ \hline
		draga & najdra\v{z}ja & predraga \\
		expensive & most expensive & too expensive \\ \hline
		nizek & najni\v{z}ji & prenizek \\
		low & lowest & too low \\ \hline
		po\v{c}asen & najpo\v{c}asnej\v{s}i & prepo\v{c}asen \\
		slow & slowest & too slow \\ \hline
		lep & najlep\v{s}i & \v{c}udovit \\
		beautiful & most beautiful & wonderful \\ \hline
        dober & najbolj\v{s}i & odli\v{c}en \\
        good & best & excellent \\
      \hline
\end{tabularx}
\caption{Examples of frequent errors in Slovenian in the ``Superlative adjective'' category. The first two columns represent words $c$ and $d$, respectively, from the analogy dataset. In the third column is the best prediction of the word from the second column, as described in Section~\ref{sec:evaluation}. Each example has an English translation in the line below the example.}
 \label{tab:error-superlative}
 \end{center}
\end{table}

Several examples of identified issues in Slovenian in the category ``Superlative adjective'' are shown in Table \ref{tab:error-superlative}.  Common issues are the adjectives with the opposite meaning, synonyms, and other forms of adjective comparison. An example of the latter is, given the base adjective "velik" (big), the superlative form is "najve\v{c}ji" (the biggest), but the predicted word is "prevelik" (too big). 

If we choose the correct prediction not only from the nearest word, but from $n$ nearest words, the evaluation scores increase in all languages in all categories. The differences between different languages then also significantly decrease. We report the scores for values of $n$ equal to 3, 5, and 10 in Tables \ref{tab:scores3}, \ref{tab:scores5}, and \ref{tab:scores10}, respectively.

\section{Conclusion}
\label{sec:conclusions}
We prepared word analogy datasets for nine languages. The datasets are suitable for evaluation of monolingual embeddings as well as cross-lingual mappings. 
We describe the choice of 15 categories, 5 semantic and 10 syntactic, and an effort to make them language and culture neutral. While the resulting datasets in nine languages are not aligned, they are nevertheless compatible enough to allow creation of cross-lingual analogy tasks for all 72 language pairs. We present basic statistics of the created datasets and their initial evaluation using fastText embeddings. The results indicate large differences across languages and categories, and show that there is a substantial room for improvement in creation of word embeddings that would better capture relations present in the language as distances in vector spaces. 

\newcite{gladkova-etal-2016-analogy} criticised that many analogy datasets are unbalanced. We have tried to improve on this issue, but further work is needed to completely mitigate it. Additionally, the dataset can be improved by adding more categories. Bigger Analogy Test Set (BATS) for English features 40 categories \cite{gladkova-etal-2016-analogy}, though many of its categories are English language specific.

As a further challenge we see creation of similar intrinsic evaluation tasks for the assessment of contextual embeddings. Such tasks would require that the existing analogies are used in sentences or other broader contexts.

The datasets of word analogy tasks for all nine languages and all language combinations are publicly available on the Clarin repository\footnote{\url{http://hdl.handle.net/11356/1261}}. 

\section{Acknowledgements}
We thank to native speakers that checked the respective monolingual analogies:
Dace Linde (Latvian) and Andrei \v{S}umakov (Russian).

The work was partially supported by the Slovenian Research Agency (ARRS) core research programme P6-0411.
This paper is supported by European Union's Horizon 2020 research and  innovation programme under grant agreement No 825153, project EMBEDDIA (Cross-Lingual Embeddings for Less-Represented Languages in European News Media).
The results of this publication reflects only the authors' view and the EU Commission is not responsible for any  use that may be made of the information it contains.


\begin{table*}[!!hp]
\begin{center}
\begin{tabular}{lrrrrrrrrr}
Category & sl & en & hr & et & fi & lv & lt & sv & ru \\ \hline 
Capitals and countries & 53.46 & 96.16 & 68.45 & 74.85 & 93.97 & 71.43 & 73.61 & 94.99 & 91.82  \\
Family & 61.69 & 98.73 & 60.06 & 67.13 & 81.78 & 59.57 & 72.61 & 78.33 & 76.85  \\
City in country & 73.57 & 94.21 & 80.39 & 75.31 & 97.16 & 82.63 & 86.51 & 95.59 & 99.23  \\
Animals & 3.16 & 42.31 & 3.19 & 3.49 & 34.01 & 4.04 & 5.78 & 23.58 & 24.58  \\
City with river & 17.60 & 58.50 & 10.14 & 18.92 & 18.44 & 5.72 & 10.98 & 45.02 & 19.33  \\
Adjective to adverb & 55.17 & 41.11 & 52.07 & 63.67 & 60.83 & 67.59 & 85.98 & 93.02 & 41.63  \\
Opposite adjective & 44.17 & 60.00 & 43.54 & 46.67 & 33.16 & 46.97 & 70.58 & 29.12 & 5.00  \\
Comparative adjective & 54.25 & 99.89 & 55.81 & 89.46 & 84.07 & 82.14 & 80.33 & 93.23 & 55.17  \\
Superlative adjective & 33.04 & 99.68 & 33.54 & 45.03 & 69.57 & 33.04 & 70.57 & 53.08 & 27.64  \\
Verb to verbal noun & 76.89 & 87.63 & 76.33 & 96.49 & 86.67 & 76.36 & 73.07 & 49.70 & 23.02  \\
Country to nationality & 51.43 & 79.89 & 66.90 & 64.93 & 59.14 & 55.52 & 67.41 & 75.86 & 71.79  \\
Singular to plural & 55.75 & 95.86 & 54.39 & 87.92 & 91.89 & 61.07 & 69.61 & 81.21 & 81.31  \\
Genitive to dative & 43.10 & N/A & 54.03 & 70.07 & 58.59 & 55.98 & 34.55 & N/A & 51.47  \\
Present to past & 68.12 & 97.50 & 80.58 & 97.83 & 92.49 & 87.32 & 82.64 & 91.67 & 90.50  \\
Present to other tense & 71.74 & 59.09 & 69.62 & 85.17 & 93.60 & 74.12 & 70.24 & 93.06 & 92.50  \\
\hline
\end{tabular}
\caption{FastText evaluation scores in \% of correctly predicted relation pairs, i.e. how often was the vector $d$ among the 3 closest vectors to the vector $b-a+c$, given a relation pair $a:b~ \approx ~c:d$.}
\label{tab:scores3}
\end{center}
\end{table*}

\begin{table*}[!!hp]
\begin{center}
\begin{tabular}{lrrrrrrrrr}
Category & sl & en & hr & et & fi & lv & lt & sv & ru \\ \hline 
Capitals and countries & 68.14 & 96.72 & 81.51 & 84.46 & 96.50 & 82.94 & 80.90 & 96.25 & 94.34  \\
Family & 69.85 & 99.64 & 65.94 & 75.09 & 86.67 & 65.79 & 77.17 & 85.71 & 81.17  \\
City in country & 86.42 & 95.57 & 90.93 & 83.85 & 98.77 & 95.07 & 93.43 & 96.82 & 99.61  \\
Animals & 6.77 & 55.00 & 6.06 & 7.62 & 43.02 & 7.02 & 10.71 & 31.61 & 31.49  \\
City with river & 27.10 & 62.54 & 14.53 & 28.04 & 25.06 & 8.69 & 14.43 & 49.69 & 22.06  \\
Adjective to adverb & 62.19 & 48.81 & 60.21 & 67.24 & 64.83 & 75.52 & 91.53 & 95.87 & 50.49  \\
Opposite adjective & 47.92 & 63.86 & 47.37 & 51.23 & 36.32 & 49.62 & 72.69 & 34.39 & 5.00  \\
Comparative adjective & 65.63 & 100.00 & 63.86 & 91.17 & 85.71 & 84.52 & 86.09 & 94.95 & 60.03  \\
Superlative adjective & 42.17 & 100.00 & 39.97 & 54.97 & 70.96 & 38.84 & 79.60 & 57.14 & 31.05  \\
Verb to verbal noun & 79.11 & 89.21 & 81.07 & 97.08 & 87.50 & 79.09 & 76.47 & 56.85 & 26.19  \\
Country to nationality & 61.90 & 86.67 & 72.86 & 71.60 & 59.83 & 60.06 & 73.09 & 77.14 & 71.94  \\
Singular to plural & 68.64 & 97.44 & 63.02 & 90.53 & 91.97 & 68.22 & 77.50 & 85.82 & 87.11  \\
Genitive to dative & 51.77 & N/A & 64.62 & 71.68 & 61.85 & 63.93 & 40.78 & N/A & 58.51  \\
Present to past & 78.80 & 99.33 & 86.78 & 99.67 & 93.87 & 90.40 & 85.76 & 91.67 & 93.00  \\
Present to other tense & 80.43 & 71.27 & 74.88 & 90.00 & 95.04 & 76.40 & 72.82 & 95.31 & 93.75  \\
\hline
\end{tabular}
\caption{FastText evaluation scores in \% of correctly predicted relation pairs, i.e. how often was the vector $d$ among the 5 closest vectors to the vector $b-a+c$, given a relation pair $a:b~ \approx ~c:d$.}
\label{tab:scores5}
\end{center}
\end{table*}

\begin{table*}[!!hp]
\begin{center}
\begin{tabular}{lrrrrrrrrr}
Category & sl & en & hr & et & fi & lv & lt & sv & ru \\ \hline 
Capitals and countries & 80.67 & 98.00 & 90.75 & 91.67 & 97.94 & 90.19 & 87.25 & 97.95 & 96.97  \\
Family & 76.92 & 99.82 & 73.07 & 81.31 & 91.11 & 72.25 & 83.26 & 90.71 & 87.35  \\
City in country & 94.64 & 97.59 & 95.85 & 91.06 & 99.62 & 98.95 & 97.50 & 97.70 & 99.72  \\
Animals & 11.63 & 70.60 & 11.06 & 16.93 & 52.93 & 14.21 & 22.37 & 44.56 & 40.09  \\
City with river & 38.16 & 68.16 & 22.80 & 42.23 & 30.02 & 12.50 & 19.11 & 58.10 & 26.05  \\
Adjective to adverb & 70.75 & 60.21 & 69.82 & 72.91 & 69.52 & 82.87 & 96.69 & 98.15 & 61.58  \\
Opposite adjective & 51.67 & 68.18 & 49.28 & 57.19 & 37.89 & 50.00 & 72.88 & 43.86 & 6.67  \\
Comparative adjective & 74.14 & 100.00 & 71.01 & 93.30 & 87.36 & 85.71 & 89.20 & 95.69 & 67.43  \\
Superlative adjective & 52.17 & 100.00 & 48.43 & 66.08 & 71.74 & 44.87 & 83.44 & 64.12 & 35.61  \\
Verb to verbal noun & 81.33 & 92.89 & 84.02 & 98.83 & 87.50 & 81.14 & 78.95 & 63.99 & 26.98  \\
Country to nationality & 72.38 & 92.26 & 79.29 & 76.93 & 61.21 & 66.23 & 78.15 & 79.86 & 71.94  \\
Singular to plural & 79.68 & 97.44 & 73.35 & 93.14 & 92.04 & 77.60 & 86.19 & 89.33 & 92.04  \\
Genitive to dative & 60.10 & N/A & 73.95 & 73.75 & 65.36 & 71.79 & 47.00 & N/A & 64.92  \\
Present to past & 82.61 & 99.83 & 92.36 & 100.00 & 95.45 & 93.48 & 91.49 & 91.67 & 94.25  \\
Present to other tense & 85.51 & 74.91 & 78.71 & 95.50 & 95.45 & 79.71 & 74.80 & 95.83 & 95.00  \\
\hline
\end{tabular}
\caption{FastText evaluation scores in \% of correctly predicted relation pairs, i.e. how often was the vector $d$ among the 10 closest vectors to the vector $b-a+c$, given a relation pair $a:b~ \approx ~c:d$.}
\label{tab:scores10}
\end{center}
\end{table*}
\section{Bibliographical References}
\label{main:ref}

\bibliographystyle{lrec}
\bibliography{lrec2020W-analogies}


\end{document}